# Hyper-Fusion Network for Semi-Automatic Segmentation of Skin Lesions

Lei Bi[a, *], Michael Fulham[a, b], and Jinman Kim[a, *]

[a] School of Computer Science, University of Sydney, NSW, Australia

[b] Department of Molecular Imaging, Royal Prince Alfred Hospital, NSW, Australia

[*] Corresponding authors: lei.bi@sydney.edu.au, jinman.kim@sydney.edu.au

*Abstract* — Segmentation of skin lesions is an important step for imaging-based clinical decision support systems. Automatic skin lesion segmentation methods based on fully convolutional networks (FCNs) are regarded as the state-of-the-art for accuracy. When there are, however, insufficient training data to cover all the variations in skin lesions, where lesions from different patients may have major differences in size/shape/texture, these methods failed to segment the lesions that have image characteristics, which are less common in the training datasets. FCN-based semi-automatic segmentation methods, which fuse user-inputs with high-level semantic image features derived from FCNs offer an ideal complement to overcome limitations of automatic segmentation methods. These semi-automatic methods rely on the automated state-of-the-art FCNs coupled with user-inputs for refinements, and therefore being able to tackle challenging skin lesions. However, there are a limited number of FCN-based semi-automatic segmentation methods and all these methods focused on 'early-fusion', where the first few convolutional layers are used to fuse image features and user-inputs and then derive fused image features for segmentation. For early-fusion based methods, because the user-input information can be lost after the first few convolutional layers, consequently, the user-input information will have limited guidance and constraint in segmenting the challenging skin lesions with inhomogeneous textures and fuzzy boundaries. Hence, in this work, we introduce a hyper-fusion network (HFN) to fuse the extracted user-inputs and image features over multiple stages. We separately extract complementary features which then allows for an iterative use of user-inputs along all the fusion stages to refine the segmentation. We evaluated our HFN on three well-established public benchmark datasets

– ISBI Skin Lesion Challenge 2017, 2016 and PH2 – and our results show that the HFN is more accurate and generalizable than the state-of-the-art methods, in particular with challenging skin lesions.

## I.  INTRODUCTION

Melanoma (also known as malignant melanoma) has one of the most rapidly increasing incidences in the world and has considerable mortality rate if left untreated (Rigel et al. 1996). Early diagnosis is particularly important because melanoma can be cured with early excision (Celebi et al. 2007, Capdehourat et al. 2011). Skin lesion images such as dermoscopy are commonly acquired as a non-invasive imaging technique for the in-vivo evaluation of pigmented skin lesions and play an important role in early diagnosis (Celebi et al. 2007). The identification of melanoma from skin lesion images using human vision alone, can be subjective, inaccurate and poorly reproducible, even among experienced dermatologists (Celebi et al. 2008, Abbas et al. 2013). This is attributed to the challenges in interpreting skin lesion images where there can be diverse visual characteristics such as variations in size, shape boundaries (e.g., 'fuzzy'), artifacts and has hairs (Figure 1) (Barata et al. 2015). Therefore, automated image analysis is a valuable aid for clinical decision support (CDS) systems and for the image-based diagnosis of skin lesions (Serrano and Acha 2009, Esteva et al. 2017). Skin lesion segmentation is the fundamental for these CDS systems and has motivated the development of numerous segmentation methods.

Traditional fully automatic segmentation methods mainly focus on extracting pixel-level or region-level features such as Gaussian (Wighton et al. 2011) and texture (He and Xie 2012) features and then use various classifiers, such as Wavelet network (Sadri et al. 2012) and Bayes classifier (Wighton et al. 2011), to separate the skin lesions from surrounding healthy skin. However, their performance depends heavily on correctly tuning a large number of parameters and effective pre- and post-processing techniques such as hair removal and corrections to illumination. These methods, without pre- and post-processing techniques, have difficulty in segmenting lesions when there are artifacts, hair or when the lesion reaches the boundary of the image.

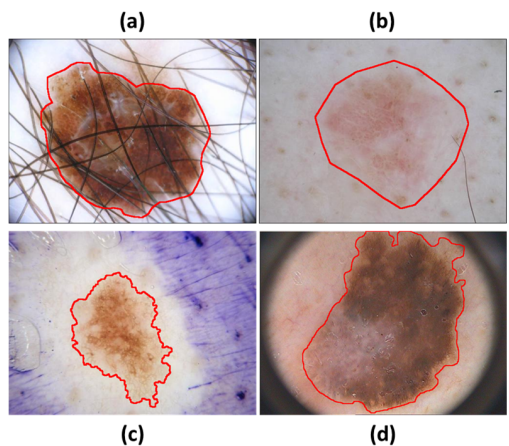

Figure 1. Examples of skin lesions with hair (a), poor contrast (b), inhomogeneity (c, d), different fields-of-view (d), and artifacts (c).

Deep learning based fully automatic segmentation methods are regarded as the state-of-the-art in skin lesion segmentation, and most of these methods are based on fully convolutional networks (FCNs) (Shelhamer et al. 2016). The success of FCNs is primarily attributed to their use of encoders and decoders to derive an image feature representation that combines low-level appearance information with high-level semantic information. The encoders use convolutional layers and downsampling processes to extract high-level semantic features from images. The decoders then upsample the extracted image features to output the segmentation results. Therefore, FCNs can be trained in an end-to-end manner for efficient inference, i.e., images are taken as inputs and the segmentation results are outputted directly. Yuan et al. (Yuan et al. 2017) replaced the cross-entropy loss used in a traditional FCN with a Jaccard distance loss for skin lesion segmentation. Yu et al. (Lequan et al. 2017) increased the FCN network depth (number of layers) with a 50-layer deep residual network for the segmentation based on deeper image features. Bi et al. (Bi et al. 2017) proposed a class-specific learning to combine (ensemble) multiple trained FCNs (trained only with melanoma or non-melanoma images) for segmentation. More recently, Xie et al. (Xie et al. 2020) proposed learning skin lesion segmentation and classification (melanoma vs. non-melanoma) via a mutual bootstrapping network, where skin lesion classification results were used to guide and improve the segmentation results. However, all these FCN-based methods are reliant on large annotated training data that include all the possible variations in skin lesions, including differences between patients in lesion size, shape and texture.

When there are, however, insufficient training data to cover all the variations in skin lesions, these methods failed to segment the lesions that have image characteristics, which are less common in the training datasets. Further, skin lesions from different datasets may have major differences in appearance e.g., illumination and field-of-view (as shown in Figure 1). The end result is that these methods tend to overfit to one dataset and have limited generalizability to a different dataset.

FCN-based semi-automatic segmentation methods for medical images, which combine manual user-inputs (priori knowledge) with high-level semantic features derived from FCNs, offer an alternative approach to segment the skin lesions. Currently, there are few such methods. Wang et al. (Wang et al. 2018) proposed a semi-automatic medical image segmentation method with two FCNs: the first FCN automatically segmented the input image, and the second FCN repeated the segmentation but with the fusion of the input image, the segmentation result (from the first FCN) and the user-inputs. The regions that failed to be segmented by the first FCN were then refined by the second FCN. Lei et al. (Lei et al. 2019) replaced the FCNs in the approach reported by Wang et al. (Wang et al. 2018), with a lightweight network architecture to segment organs-at-risk structures from computed tomography (CT) images. Koohbanani et al. (Koohbanani et al. 2020) fused user-inputs with a multi-scale FCN for microscopy images. Sakinis et al. (Sakinis et al. 2019) fused user-inputs with a U-Net for organs segmentation in abdominal CT images. Wang et al. (Wang et al. 2018) proposed fine-tuning the individual test images with user-inputs (including scribbles and user-defined bounding boxes) to enclose the regions of interest. Sun et al. (Sun et al. 2018) proposed a patch-based segmentation method where a user-defined centroid point was used to partition the medical image into small patches and the small patches were then segmented with a convolutional recurrent neural network (ConvRNN). For non-medical images, Majumder et al. (Majumder and Yao 2019) fused superpixel-based user-inputs with input images for natural image segmentation and the superpixel-based user-inputs were derived by calculating the Euclidean distance from the centroid of the superpixel to the user-clicks. All these FCN-based methods focused on early-fusion, where the medical images are fused with the user-inputs (both foreground and background inputs) as a single input prior to the FCN. The reliance on a single fused input means that the important user-input information could be lost after early-fusion, and so there will be limited priori knowledge that can be used by the FCN. In addition, the reliance on a user-defined centroid point is not always feasible. It is challenging to accurately place a centroid point for lesions with differing shapes and the centroid point may not always be within the lesion region. Further, fine-tuning individual test images requires additional computational time and manual input e.g., bounding boxes and scribbles, and this is challenging to implement for a large cohort study. Hu et al. (Hu et al. 2019) proposed a two-stream late fusion network for natural image segmentation, where the image and the user-inputs were separately processed by two FCN networks with fusion of the resultant features. The late fusion of extracted image features, however, tends to dismiss the correlations between the image and the user-inputs; the correlations may only accessible at the early stage of the network. In addition, when these methods are applied to skin lesion segmentation, they usually have difficulty in accurately delineating the boundary of the lesion and have inconsistent outcomes for the challenging skin lesions.

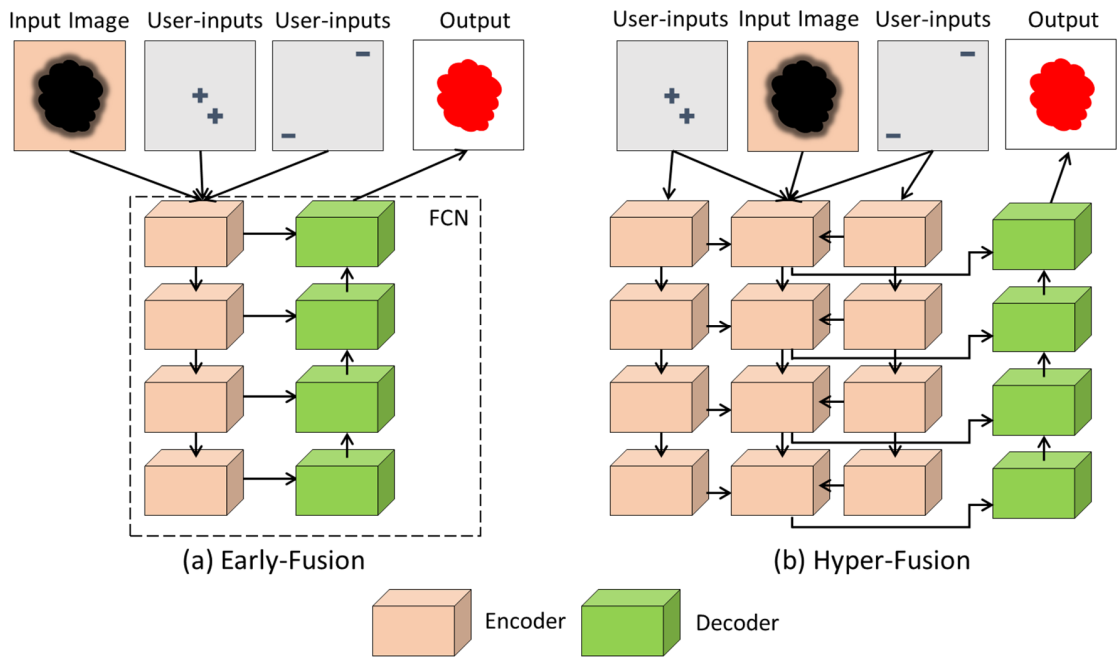

Figure 2. Different architectures for semi-automatic skin lesion segmentation. (a) methods based on early-fusion, where medical images are fused with user-inputs (foreground and background inputs) as a single input before entering the FCN; and (b) our proposed hyper-fusion network (HFN) where user-inputs are separately processed for feature extraction and then fused across multiple stages to progressively fuse intermediary semantic features with user-inputs. User-input '+' represents the foreground and '–' represents the background.

*A. Our Contribution*

Our hyper-fusion network (HFN) shown in Figure 2b, separately extracts features from user-inputs. Our fusion strategy provides the flexibility to learn complementary features between the lesion images and the user-input, and provide continuous guidance and constraint to the segmentation results. Our HFN adds the following contributions to the current knowledge: (i) separate extraction of features from skin lesion images and user-inputs; they will allow to continuous leverage of user-inputs to optimize learning of skin lesion characteristics and minimize the loss of user-input information during early-fusion. (ii) training and predicting segmentation results in multiple fusion stages. When compared to early-fusion based semi-automatic segmentation methods, multiple fusion stages have the advantage of using the user-inputs to iteratively refine the segmentation, which ensures better segmentation of skin lesion boundaries. (iii) the introduction of hyper-integration modules (HIMs) to fuse user-input features and skin lesion image features at individual fusion stages. HIMs help guide and constrain the learning of the lesion characteristics and then propagate the intermediary segmentation results to the next stage of the decoder. The fusion from individual stages ensures the appearance of the segmented skin lesions is spatially consistent.

## II. Methods and Materials

*A. Materials*

We used three well-established public benchmark datasets to train and test the effectiveness of our method.

- The 2017 and 2016 ISBI Skin Lesion Challenge (denoted as ISBI 2017 (Codella et al. 2017) and ISBI 2016 (Gutman et al. 2016)) datasets are a subset of the large International Skin Imaging Collaboration (ISIC) archive, which contains skin lesion images acquired on a variety of different devices at numerous leading international clinical centers. The ISBI 2017 challenge dataset provides 2,000 training images (1,626 non-melanoma and 374 melanoma) and a separate test dataset of 600 images (483 non-melanoma and 117 melanoma). Image size varies from 453×679 pixels to 4499×6748 pixels. The ISBI 2016 challenge dataset provides 900 training images (727 non-melanoma and 173 melanoma) and a separate test dataset of 379 images (304 non-melanoma and 75 melanoma). Image size varies from 566×679

pixels to 2848×4288 pixels.

- The PH2 public dataset (Mendonça et al. 2013) composes of 200 dermoscopic images that were collected by the Universidade do Porto, Técnico Lisboa, and the Dermatology service of Hospital Pedro Hispano in Matosinhos, Portugal. All 200 images (160 non-melanoma and 40 melanoma) were obtained under the same conditions using a Tuebinger Mole Analyzer system using a 20-fold magnification. Image size varies from 553×763 pixels to 577×769 pixels.

All datasets provided ground truth segmentations based on manual delineations by clinical experts.

### B. Hyper-Fusion Network (HFN)

Our HFN contains hyper-fusion encoder (HFE) and decoder (HFD), as shown in Figure 3. The HFE has three branches and uses downsampling processes to extract high-level semantic features from the skin lesion images, foreground and background hint maps (derived from user-inputs). The extracted intermediary features (feature maps) of the two hint map branches at each stage were further processed and fused with the skin lesion image branch. The HFD has multiple hyper-integration modules (HIMs) that upsample the feature maps derived from the encoder, with additional constraint and guidance from hint maps to output the segmentation results.

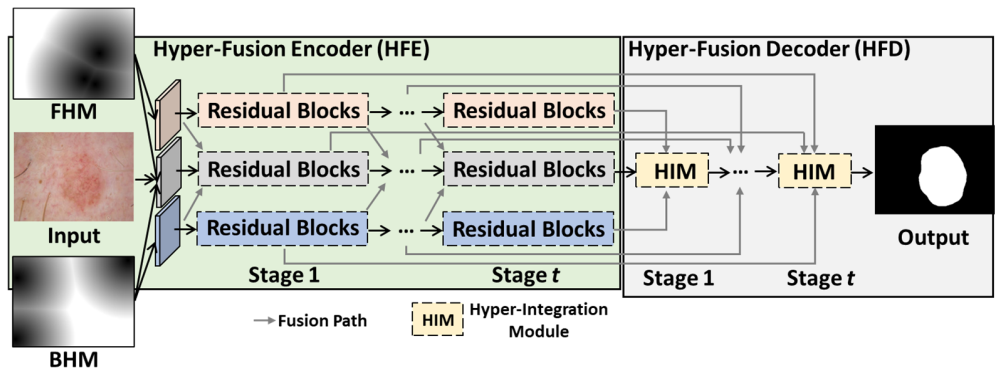

Figure 3. Flow diagram of our HFN with the input of the foreground hint map (FHM) and the background hint map (BHM).

*1) Foreground and Background Hint Maps*

Based on user-inputs (clicks), we created foreground and background hint maps to encode the probability of a pixel being within the skin lesion regions and the non-skin lesion regions (shown in Figure 4). The calculation of these two hint maps was achieved by applying the Euclidean distance transform to derive the shortest distance from a pixel to the user clicked pixels, which can be defined as:

$$f(p) \coloneqq min\{\varepsilon(p,q)|q \in \delta(f)\} \tag{1}$$

$$\beta(p) \coloneqq min\{\varepsilon(p,q)|q \in \delta(\beta)\} \tag{2}$$

where $f(p)$ and $\beta(p)$ represent pixels $p$ in a foreground hint map and in a background hint. $\varepsilon$ is the Euclidean distance and $\delta(f)$ and $\delta(\beta)$ are user clicked pixels in the foreground (skin lesion regions) and background (non-skin lesion regions).

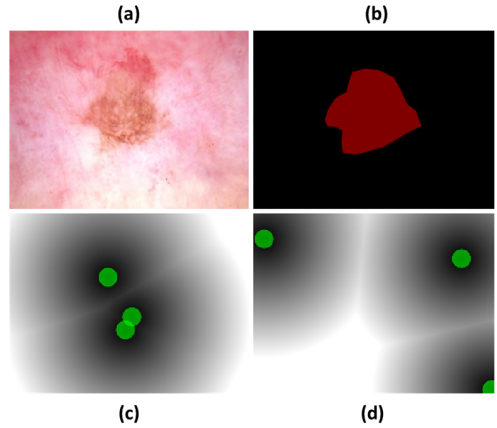

Figure 4. Foreground and background hint maps. (a) input skin lesion image; (b) ground truth annotation; (c) foreground hint map; and (d) background hint map. Green circles indicate user clicked pixels (user-inputs).

*2) Hyper-Fusion Encoder (HFE)*

The HFE was based on the residual neural network (ResNet) (He et al. 2016), given to its popularity for various skin lesion related tasks (Lequan et al. 2017). Our HFE has multiple stages ($t$), which represent different-sized output feature maps. Each stage consists of a number of residual blocks and each residual block has a skip connection that bypasses (shortcuts) a few convolutional (Conv), batch normalization (BN) and ReLU layers at a time. The input feature maps at each stage were initially fused with the output feature maps from the other branches, as shown in Figure 3. A SUM layer and ReLU layer were used for this fusion according to:

$$HFE^t(I) = \varphi^t(ReLU(SUM(HFE^{t-1}(I), HFE^{t-1}(f), HFE^{t-1}(\beta)))) \qquad (3)$$

where $I$ is the input skin lesion image, $f$ and $\beta$ are the foreground and background hint maps, and $\varphi^t$ represents the network operations e.g., Conv, at stage $t$.

*3) Hyper-Fusion Decoder (HFD)*

In the HFD, we built multiple hyper-integration modules (HIMs) to fuse and skip-connect the output features from skin lesion image branches and from foreground and background hint map branches. The HIM consisting of a guidance and constraint unit (GCU), a fusion unit (FU) and a chained residual pooling unit (CRPU), as shown in Figure 5.

As for the guidance and constraint unit (GCU), the design was based on a convolutional block attention module (CBAM) (Woo et al. 2018). When compare to CBAM, GCU differs in that it focuses on extracting guidance and constraint information across the foreground and the background hint maps, to be used by the skin lesion image feature maps. We placed the GCU at the start of each HIM to ensure different input feature maps can be learned collaboratively for segmentation. The first half of the GCU was designed to explore channel-wise context and we define this as the inter-channel relationship within the feature maps of the input foreground and background hint maps. After the feature maps were concatenated, we used max pooling (Max Pool) and average pooling (Avg Pool) on the input multi-modality feature to encode and aggregate global information for each channel. After that, two fully connected layers were inter-connected via ReLU layer.

Our motivation for this design was to activate the neurons that were related to both feature maps. The output feature maps were summed and passed to a Sigmoid layer to produce a smoothed output. The second half of the GCU was designed to identify spatial context, which are the spatial locations for the network to focus on. We concatenated the output features from the max pooling and average pooling layers. We applied convolution, batch normalization (BN), ReLU and Sigmoid operations to encode the spatial locations on the concatenated feature maps. At the end of GCU, the locations were multiplied with the output feature maps derived from the skin lesion images.

The fusion unit (FU) fuses the output feature maps from the previous HFD stage and outputs a higher-resolution feature map (as shown in Figure 5). The output feature maps $HFD^{t-1}$ was firstly upsampled via linear interpolation to have the same dimension to the current stage $t$ feature map. The upsampled feature maps were then summed with the intermediary feature maps at the current stage.

We used the chained residual pooling unit (CRPU) at the end of HIM to refine the segmentation results (Lin et al. 2019). As shown in Figure 5, CRPU consists of multiple max pooling layers and were fused iteratively. Therefore, the CRPU module allows the encoding of contextual information and the refinement of the segmentation results from different size pooled regions. In our implementation, we iteratively pooled the input feature maps 4 times.

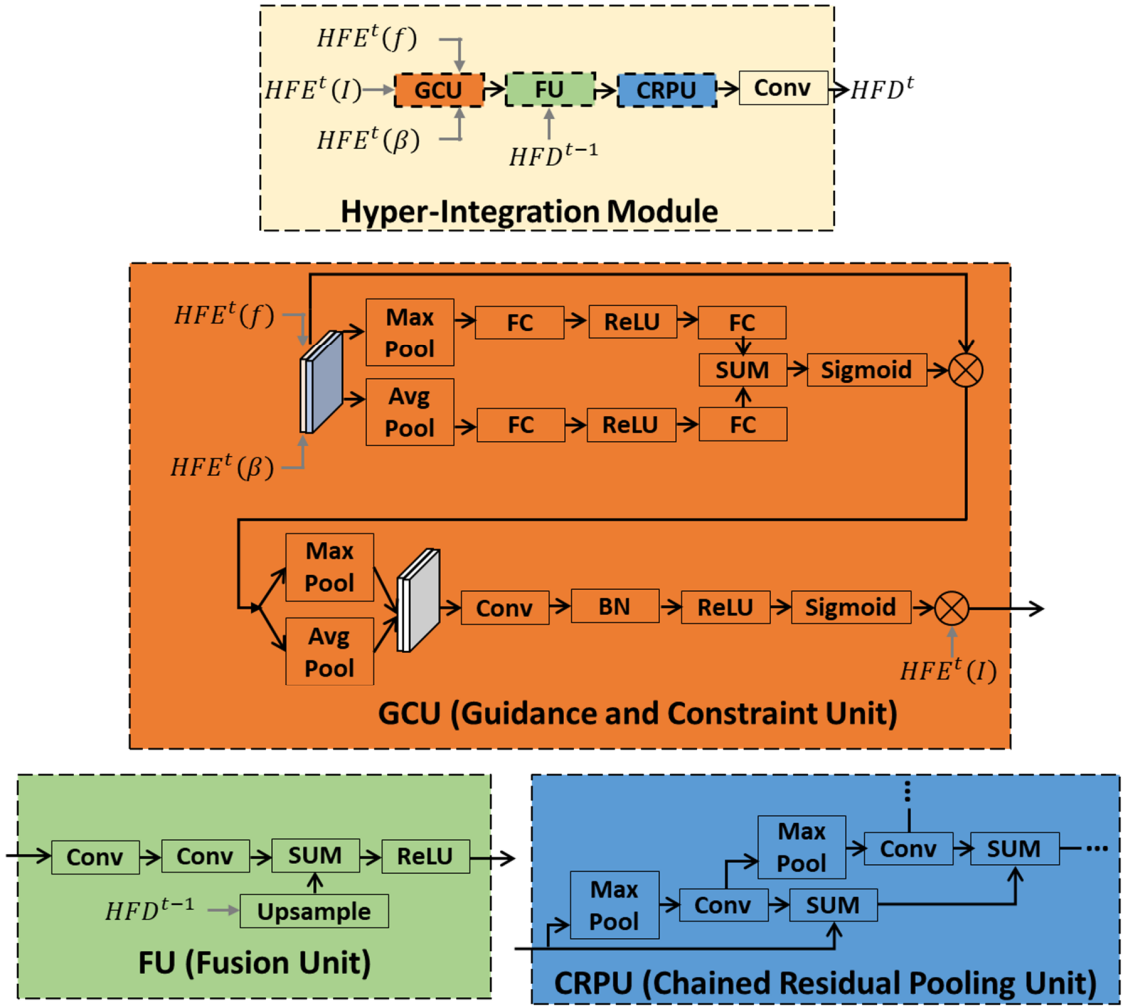

Figure 5. Proposed Hyper-Integration Module (HIM).

*4)  Simulated User-Inputs*

Similar to the approach proposed by Wang et al. (Wang et al. 2018), we simulated user-inputs (user clicked pixels) to derive the foreground and background hint maps for each skin lesion image. We used the ground truth annotations to randomly sample skin lesion pixels $\delta(f)$ and non-skin lesion pixels $\delta(\beta)$. To ensure the randomly sampled pixels are apart from each other (to avoid cluttered pixels), we fragmented the ground truth annotations into disjoint sets, where all the pixels in a set will have similar Euclidean distances to the skin lesion boundaries. For each set, we only sampled one user clicked pixel.

During the training stage, for each skin lesion image, we sampled 6 different combinations of $\delta(f)$ and $\delta(\beta)$, where $\delta(f)$ and $\delta(\beta)$ consist of 1 to 6 pixels (accumulated). This upper limit (6 different combinations) was empirically selected, where we found that the segmentation results tend to converge before the upper limit. In each combination, the number of $\delta(f)$ and $\delta(\beta)$ is the same, as shown in Figure 6.

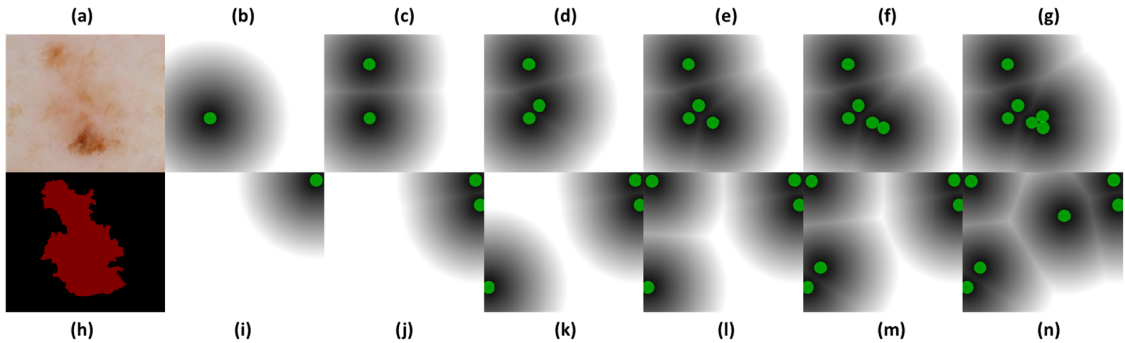

Figure 6. Examples of simulated user-inputs to a melanoma image. (a) input image with inhomogeneous textures and fuzzy boundaries; (h) ground truth; (b-g) foreground hint maps; and (i-n) background hint maps. The foreground and background hint maps were derived when $\delta(f)$ and $\delta(\beta)$ were set to between 1 to 6 user clicked pixels.

*5) Implementation and Training Details*

For skin lesion segmentation, our HFN was trained end-to-end by minimizing the overall loss between the predicted results and the ground truth annotations of the training data:

$$arg\min_{\theta} \sum \mathcal{L}(Y, Z|\theta) \qquad (4)$$

where $\mathcal{L}$ measures the cross-entropy loss of the expert-provided annotation $Z$ and the predicted results $Y$. The network parameters ($\theta$) was updated iteratively using the stochastic gradient descent (SGD) (Dean et al. 2012) algorithm.

Similar to published data (Shin et al. 2016), we used the pre-trained 101-layer-ResNet (trained on ImageNet (Russakovsky et al. 2015)) network for initialization. 101-layer-ResNet consists of 4 stages ($t = 4$) and had 3, 4, 23 and 3 residual blocks (bottleneck block) at each stage. In addition, given the marked variation in the size of skin lesion images, we pre-processed and resized (nearest-neighbor interpolation) all the training skin lesion images to be at a maximum of 512 pixels in the longest axis while preserving image's aspect ratios. We used data augmentation techniques, including random cropping and flipping, to improve the generalizability of training. The HFN was implemented with PyTorch and was fine-tuned for 150 epochs with a batch size of 3 on a 11GB 2080Ti GPU. The initial learning rate was set to 0.0005 for HFE and 0.005 for HFD. The learning rate was halved after every 50 epochs. The momentum was set to 0.9 and weight decay was set to 0.00001.

## C. Experiment Setup

We performed the following experiments on the three datasets: (a) comparison of the overall performance of our method with fully automated and semi-automated segmentation methods; (b) comparison of the results from (a) using different number of user-inputs; (c) analysis of the performance of each component in our proposed method; (d) analysis of the segmentation results on the challenging skin lesions; and (e) analysis of the segmentation results with noisy user-inputs. For experiments using the ISBI 2017 and ISBI 2016 datasets, we used the specified training and test dataset. For experiments on the PH2 dataset, we followed the protocol used by Yuan et al. (Yuan et al. 2017): training on the ISBI 2016 training dataset and tested on the PH2 dataset.

State-of-the-art automated and semi-automated skin lesion segmentation methods used in the comparison. The automated skin lesion segmentation methods include: (1) the top 5 results out of 21 teams for the ISBI 2017 challenge (Codella et al. 2017); (2) the top 5 out of 28 teams for the ISBI 2016 challenge (Gutman et al. 2016); (3) DT (Pennisi et al. 2016) – skin lesion segmentation using Delaunay triangulation; (4) SCDRR (Bozorgtabar et al. 2016) – sparse coding with dynamic rule-based refinement; (5) JCLMM (Roy et al. 2017) – skin lesion segmentation by joining circular-linear distributions with mixture models; (6) FCN-8s (Shelhamer et al. 2016) – visual geometry group network (VGGNet) based fully convolutional networks (at stride size 8); (7) MFCN (Bi et al. 2017) – skin lesion image segmentation via multi-stage fully convolutional networks; (8) J-FCN (Yuan et al. 2017) – a Jaccard distance based fully convolutional network for skin lesion segmentation; (9) DCL (Bi et al. 2019) – deep class-specific learning which learns the important visual characteristics of the skin lesions of each individual class (melanoma vs non-melanoma) on an individual basis; (10) MB (Xie et al. 2020) – jointly learn skin lesion segmentation and classification via a mutual bootstrapping network; (11) FrCN (Al-Masni et al. 2018) – full resolution convolutional networks; (12)

BiDFL (Wang et al. 2019) – a bi-directional dermoscopic feature learning with multi-scale consistency analysis for skin lesions segmentation; (13) DAGAN (Lei et al. 2020) – generative adversarial network with dual discriminators; and (14) DDN (Li et al. 2018) – skin lesion segmentation with dense deconvolutional network; and (15) ENS (Goyal et al. 2019) – an ensemble of results derived from Mask R-CNN (He et al. 2017) and DeepLabV3+ (Chen et al. 2018). The included semi-automatic segmentation methods are: (1) FCNN (Sakinis et al. 2019) – an early-fusion approach with U-Net for segmentation; (2) CAGN (Majumder and Yao 2019) – content-aware guidance map network, where superpixel-based user-inputs were derived by calculating the Euclidean distance from the centroid of the superpixel to the user-clicks; (3) FCTSFN (Hu et al. 2019) – fully convolutional two-stream fusion network, where the images and the user-inputs were separately processed by two FCN networks with the resultant features fused; and (4) P-Net (Wang et al. 2018) – a semi-automatic FCN based segmentation method based on early fusion, where all the feature maps from the encoder were concatenated for decoding and segmentation. We replaced VGGNet backbone used in P-Net with a 101-layer-ResNet backbone for fair comparison.

When comparing to automatic segmentation methods, we set both the $\delta(f)$ and the $\delta(\beta)$ equal to (3, 3) for all the semi-automatic segmentation methods; these setting was based on our empirical results where we varied the user-input from (1, 1) to (6, 6) with the results stabilizing from (3, 3). The comparison on using different number of user-inputs were evaluated in a separate experiment. All the semi-automatic segmentation methods also used the same simulated user-inputs.

For the analysis of the segmentation results on the challenging skin lesions, these skin lesions were selected based on the segmentation results derived from the conventional FCN-8s; segmentation results ranked in the bottom 20% were considered as the challenging skin lesions. The conventional FCN-8s was used to be objective in the selection process.

We evaluated the segmentation results with different number of noisy foreground and background user-inputs. Noisy user-inputs were simulated by randomly selected a pixel which was 5 to 10 pixels away from the ground truth annotation. In our pre-processing, skin lesion image was resized to be at a maximum of 512 pixels in the longest axis and the average diameter of the skin lesion was ~180 pixels. Therefore, we suggest that 5 to 10 pixels away from the ground truth annotation representing ~2.8%-5.6% error is a reasonable representation of noisy user-inputs.

*D. Evaluation Metrics*

Standard skin lesion segmentation evaluation metrics were used, including the Jaccard index (Jac.), sensitivity (Sen.), specificity (Spec.) and accuracy (Acc.). In addition, we calculated the precision-recall (PR) curves, which have been widely used for object detection and segmentation problems on general images (Li et al. 2013).

## III. Experiments and Results

*A. Segmentation Results on ISBI 2017, ISBI 2016 and PH2 Datasets*

Tables 1-2 and Figure 7 show that our HFN method achieved the best overall performance across all measurements on the ISBI 2017 dataset. When compared with the recently published fully automatic methods of MB and BiDFL, our method improved by a large margin of 3.3% and 2.23% in Jaccard measure (Table 1).

Table 1. Comparison of segmentation results on ISBI 2017 dataset for all studies (non-melanoma and melanoma skin lesions). (Segmentation results have been sorted in an ascending order based on Jaccard index; * = the best results; Train = training data required; Semi = semi-automatic segmentation methods; and Deep = deep learning based segmentation methods).

| ISBI 2017 – Overall | Jac. | Sen. | Spec. | Acc. | Train | Semi | Deep |
|---|---|---|---|---|---|---|---|
| SSLS | 44.77 | 46.29 | 99.40* | 83.92 | | | |
| FCN-8s | 73.12 | 83.57 | 95.96 | 92.86 | √ | | √ |
| Team - RECOD Titans | 75.40 | 81.70 | 97.00 | 93.10 | √ | | √ |
| Team - BMIT | 75.80 | 80.10 | 98.40 | 93.40 | √ | | √ |
| Team - BMIT | 76.00 | 80.20 | 98.50 | 93.40 | √ | | √ |
| Team - NLP LOGIX | 76.20 | 82.00 | 97.80 | 93.20 | √ | | √ |
| FCNN | 76.28 | 84.47 | 96.70 | 93.49 | √ | √ | √ |
| Team - Mt.Sinai | 76.50 | 82.50 | 97.50 | 93.40 | √ | | √ |
| DDN | 76.50 | - | - | 93.90 | √ | | √ |
| DAGAN | 77.10 | 83.50 | 97.60 | 93.50 | √ | | √ |
| FrCN | 77.11 | 85.40 | 96.69 | 94.03 | √ | | √ |
| DCL | 77.73 | 86.20 | 96.71 | 94.08 | √ | | √ |
| ENS | 79.34 | 89.93 | 95.00 | 94.08 | √ | | √ |
| CAGN | 79.67 | 87.64 | 96.61 | 94.71 | √ | √ | √ |
| FCTSFN | 79.68 | 87.63 | 96.40 | 94.60 | √ | √ | √ |
| MB | 80.40 | 87.40 | 96.80 | 94.70 | √ | | √ |
| P-Net | 80.88 | 89.16 | 96.33 | 94.99 | √ | √ | √ |
| BiDFL | 81.47 | - | - | 94.65 | √ | | √ |
| HFN | 83.70* | 92.33* | 96.16 | 95.80* | √ | √ | √ |

Table 2. Comparison of segmentation results on ISBI 2017 dataset with separated non-melanoma and melanoma skin lesion studies. * = the best results

| ISBI 2017 | Non-melanoma | | | | Melanoma | | | |
|---|---|---|---|---|---|---|---|---|
| | Jac. | Sen. | Spec. | Acc. | Jac. | Sen. | Spec. | Acc. |
| **Team - Mt.Sinai** | 77.78 | 83.85 | 97.69 | 94.18 | 71.20 | 76.99 | 96.88 | 89.96 |
| **Team - NLP LOGIX** | 78.02 | 83.99 | 97.93 | 94.23 | 68.82 | 73.75 | 97.52 | 89.03 |
| **Team - BMIT** | 77.60 | 81.85 | 98.57 | 94.34 | 69.28 | 73.40 | 97.99 | 89.64 |
| **Team - BMIT** | 77.45 | 81.81 | 98.54 | 94.31 | 69.06 | 73.26 | 97.97 | 89.56 |
| **Team - RECOD Titans** | 77.00 | 83.46 | 97.04 | 94.02 | 68.78 | 74.35 | 96.66 | 89.37 |
| **SSLS** | 46.54 | 48.17 | 99.37* | 85.42 | 37.43 | 38.52 | 99.52* | 77.73 |
| **FCN-8s** | 73.74 | 84.30 | 96.30 | 93.55 | 70.55 | 80.56 | 94.54 | 90.02 |
| **DCL** | 79.07 | 87.54 | 97.35 | 95.05 | 72.18 | 80.67 | 94.07 | 90.08 |
| **P-Net** | 81.84 | 89.84 | 96.88 | 95.62 | 76.91 | 86.32 | 94.04 | 92.38 |
| **BiDFL** | 82.49 | - | - | 95.28 | 77.26 | - | - | 92.02 |
| **HFN** | 84.36* | 92.97* | 96.71 | 96.35* | 80.98* | 89.67* | 93.89 | 93.57* |

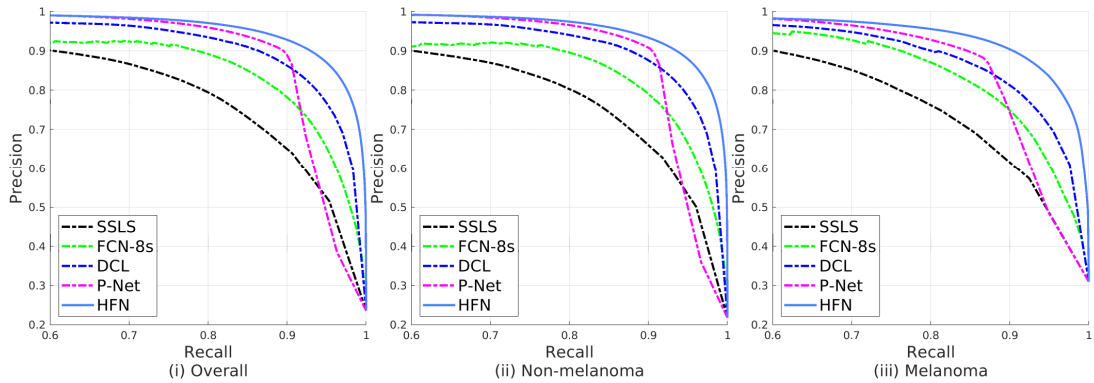

Figure 7. PR curves of different methods on ISBI2017 skin lesion challenge dataset, where (i, ii, and iii) represent overall, non-melanoma and melanoma studies.

Tables 3-4 and Figure 8 show that our HFN method outperformed all the current methods on the ISBI 2016 dataset. When compared to the current state-of-the-art method DAGAN, our method improved by 1.07% in Jaccard measure (Table 3). In addition, our HFN improved the existing methods with a large margin of 2.81% on the Jaccard measure (Table 4) for more difficult melanoma skin lesions compared to non-melanoma counterpart.

Table 3. Comparison of segmentation results on ISBI 2016 dataset for all studies.

| ISBI 2016 – Overall | Jac. | Sen. | Spec. | Acc. | Train | Semi | Deep |
|---|---|---|---|---|---|---|---|
| SSLS | 57.20 | 70.04 | 97.31 | 84.67 | | | |
| Team - TMU | 81.10 | 83.20 | 98.70* | 94.60 | √ | | √ |
| Team - SFU | 81.11 | 91.50 | 95.50 | 94.40 | √ | | √ |
| FCN-8s | 81.37 | 91.70 | 94.90 | 94.13 | √ | | √ |
| Team - Rahman | 82.22 | 88.00 | 96.90 | 95.20 | √ | | √ |
| FCNN | 82.30 | 91.72 | 95.84 | 94.69 | √ | √ | √ |
| Team - CUMED | 82.90 | 91.10 | 95.70 | 94.90 | √ | | √ |
| CAGN | 83.89 | 90.11 | 96.88 | 95.68 | √ | √ | √ |
| Team - ExB | 84.30 | 91.00 | 96.50 | 95.30 | √ | | √ |
| MFCN | 84.64 | 92.17 | 96.54 | 95.51 | √ | | √ |
| J-FCN | 84.70 | 91.80 | 96.60 | 95.50 | √ | | √ |
| P-Net | 85.83 | 93.71 | 95.49 | 95.87 | √ | √ | √ |
| DCL | 85.92 | 93.11 | 96.05 | 95.78 | √ | | √ |
| FCTSFN | 85.98 | 92.02 | 96.55 | 96.08 | √ | √ | √ |
| DDN | 87.00 | 95.10* | 96.00 | 95.90 | √ | | √ |
| DAGAN | 87.10 | 93.70 | 96.80 | 96.00 | √ | | √ |
| HFN | 88.17* | 94.22 | 96.45 | 96.64* | √ | √ | √ |

Table 4. Comparison of segmentation results on ISBI 2016 dataset for non-melanoma and melanoma studies.

| ISBI 2016 | Non-melanoma | | | | Melanoma | | | |
|---|---|---|---|---|---|---|---|---|
| | Jac. | Sen. | Spec. | Acc. | Jac. | Sen. | Spec. | Acc. |
| Team - ExB | 84.64 | 91.12 | 97.22 | 95.78 | 82.94 | 90.57 | 93.84 | 93.23 |
| Team - CUMED | 82.95 | 90.82 | 96.55 | 95.30 | 82.90 | 92.47 | 92.34 | 93.21 |
| Team - Rahman | 82.04 | 87.84 | 97.51 | 95.70 | 82.65 | 88.72 | 94.44 | 93.22 |
| Team - SFU | 80.88 | 91.55 | 95.82 | 94.93 | 81.88 | 91.16 | 94.13 | 92.19 |
| Team - TMU | 80.73 | 82.89 | 99.04* | 94.87 | 82.31 | 84.62 | 97.48* | 93.43 |
| SSLS | 58.34 | 72.87 | 97.15 | 86.15 | 52.59 | 58.58 | 97.94* | 78.67 |
| FCN-8s | 81.38 | 91.17 | 95.87 | 94.55 | 81.33 | 93.83 | 90.98 | 92.39 |
| MFCN | 84.34 | 91.63 | 97.20 | 95.71 | 85.84 | 94.34 | 93.89 | 94.70 |
| DCL | 85.60 | 92.95 | 96.79 | 96.15 | 85.62 | 93.77 | 93.05 | 94.29 |
| P-Net | 85.72 | 93.42 | 96.11 | 96.11 | 86.28 | 94.86 | 92.95 | 94.90 |
| HFN | 87.94* | 93.95* | 96.99 | 96.78* | 89.09* | 95.31* | 94.28 | 96.10* |

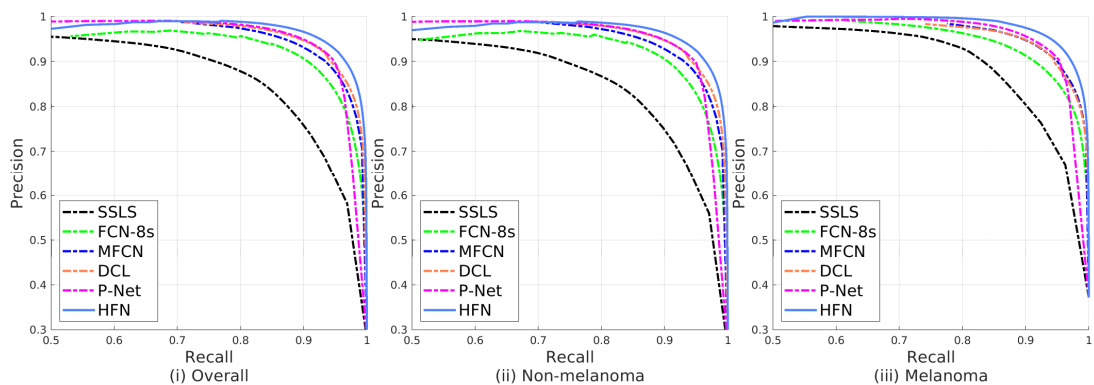

Figure 8. PR curves of different methods on ISBI2016 skin lesion challenge dataset.

We evaluate the generalizability of the proposed method and the comparison methods with the PH2 dataset. The evaluation results are presented in Tables 5-6 and Figure 9. It shows that our method achieved leading performance for both non-melanoma and melanoma studies on the PH2 dataset. It also shows that our method has a better generalizability with an improvement of 2.91% in Jaccard measure to the recently published DCL method.

Table 5. Comparison of segmentation results on PH2 dataset for all studies.

| PH2 – Overall | Jac. | Sen. | Spec. | Acc. | Train | Semi | Deep |
|---|---|---|---|---|---|---|---|
| DT | - | 80.24 | 97.22 | 89.66 | | | |
| JCLMM | - | - | - | - | √ | | |
| SSLS | 68.16 | 75.32 | 98.18* | 84.85 | | | |
| SCDRR | 76.00 | - | - | - | | | |
| FCNN | 76.35 | 95.48 | 89.48 | 91.76 | √ | √ | √ |
| FCN-8s | 82.15 | 93.14 | 93.00 | 93.48 | √ | | √ |
| FCTSFN | 83.45 | 95.64 | 94.11 | 94.17 | √ | √ | √ |
| CAGN | 83.84 | 94.26 | 95.31 | 94.03 | √ | √ | √ |
| ENS | 83.90 | 93.20 | 92.90 | 93.80 | √ | | √ |
| MFCN | 83.99 | 94.89 | 93.98 | 94.24 | √ | | √ |
| FrCN | 84.79 | 93.72 | 95.65 | 95.08 | √ | | √ |
| P-Net | 85.21 | 98.10* | 92.13 | 94.92 | √ | √ | √ |
| DCL | 85.90 | 96.23 | 94.52 | 95.30 | √ | | √ |
| HFN | 88.81* | 95.60 | 95.76 | 96.36* | √ | √ | √ |

Table 6. Comparison of segmentation results on PH2 dataset for non-melanoma and melanoma studies.

| PH2 | Non-melanoma | | | | Melanoma | | | |
|---|---|---|---|---|---|---|---|---|
| | Jac. | Sen. | Spec. | Acc. | Jac. | Sen. | Spec. | Acc. |
| DT | - | 86.79 | 97.47 | 93.74 | - | 54.04 | 95.97 | 66.15 |
| SSLS | 75.52 | 83.96 | 98.05* | 91.77 | 38.73 | 40.74 | 98.67* | 57.16 |
| FCN-8s | 82.01 | 94.83 | 94.22 | 94.79 | 82.72 | 91.39 | 88.16 | 88.25 |
| MFCN | 84.15 | 95.64 | 95.12 | 95.61 | 83.35 | 91.88 | 89.42 | 88.78 |
| DCL | 86.05 | 97.11 | 95.85 | 96.61 | 85.33 | 92.70 | 89.19 | 90.05 |
| P-Net | 84.92 | 98.73* | 93.93 | 95.88 | 86.34 | 95.56* | 84.94 | 91.09 |
| HFN | 88.75* | 96.07 | 96.74 | 97.30* | 89.02* | 93.69 | 91.87 | 92.58* |

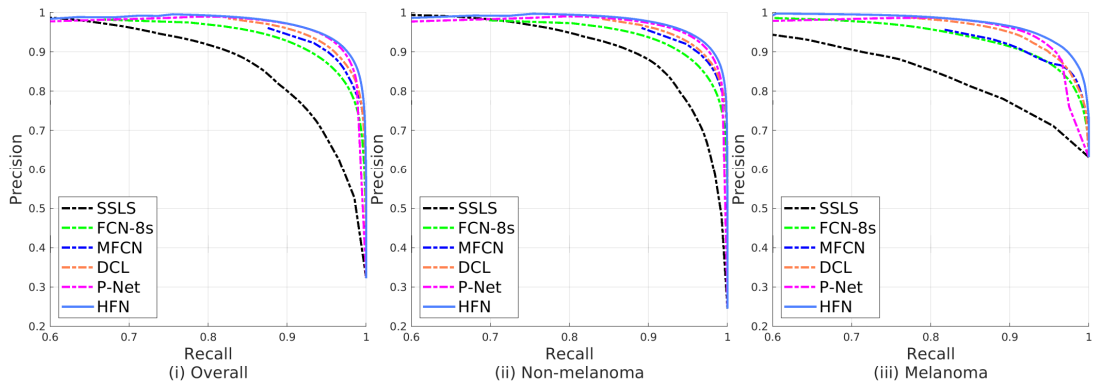

Figure 9. PR curves of different methods on the PH2 dataset.

Figure 10 shows example segmentation results, where existing automated segmentation methods failed to segment the skin lesions, which have low-contrast ratios to the background, as shown in Figure 10b, and skin lesion images acquired from a different dataset (PH2), which present different field of views, as shown in Figure 10c.

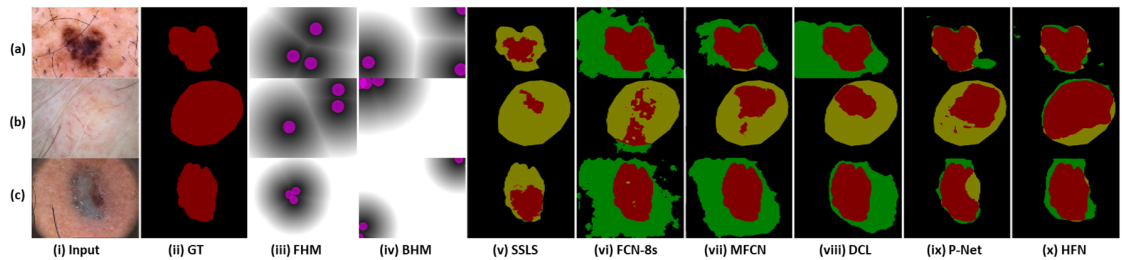

Figure 10. Segmentation results from three challenging skin lesions: (i) input images; (ii) ground truth (GT); (iii) FMH – foreground hint maps ($\delta(f) = 3$); (iv) BHM – background hint maps ($\delta(\beta) = 3$); and, (v-x) segmentations obtained by SSLS, FCN, MFCN, DCL, P-Net and our HFN. The colors represent true positive (red), true negative (black), false positive (green) and false negative (yellow) pixels. Purple circles indicate user clicked pixels.

*B. Segmentation Results with Different Number of User-Inputs*

Figure 11 shows the segmentation results with different number of user-inputs for two semi-automatic segmentation methods. As expected, the segmentation results were improved after every additional user-inputs. It also shows that our HFN consistently outperformed the P-Net method.

Figure 12 shows the segmentation results of an example skin lesion derived from HFN and P-Net. It shows that our HFN can consistently outperform P-Net across different number of user-inputs; Our HFN achieved a satisfactory segmentation result when $\delta(f)$, $\delta(\beta)$ are equal to 4. In contrast, P-Net has limited improvement with additional user-inputs and the segmentation results were still inaccurate even when $\delta(f)$, $\delta(\beta)$ were equal to 6. These results suggest that our HFN method can better leverage user-inputs.

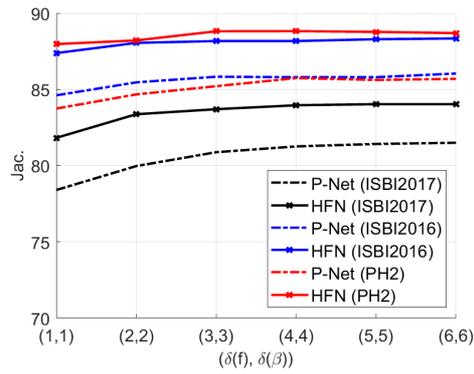

Figure 11. Overall segmentation results of HFN and P-Net with different number of user-inputs ($\delta(f)$ and $\delta(\beta)$ ).

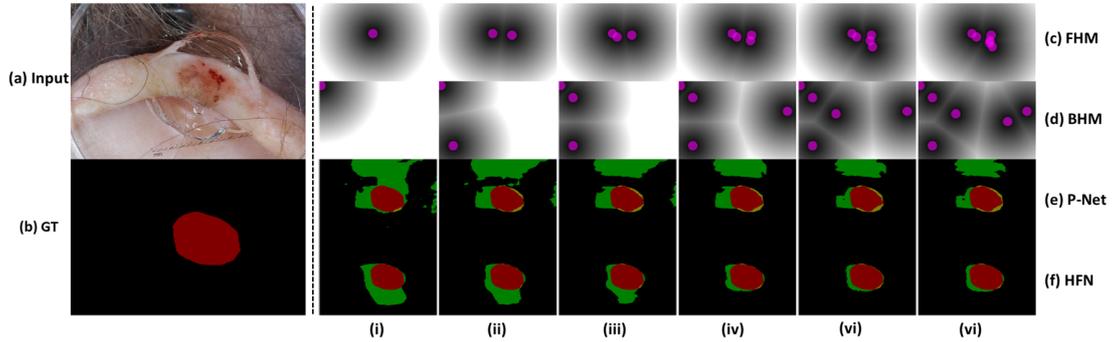

Figure 12. Example segmentation results of a skin lesion with our HFN and P-Net using different number ($\delta(f)$ and $\delta(\beta)$ ranging from 1 to 6) of user inputs (i - vi), where (a) input image; (b) ground truth (GT); (c) foreground hint maps; (d) background hint maps; (e, f) segmentation results derived from P-Net and HFN.

*C. HFN Component Analysis*

Segmentation results of our HFN with and without HIMs are presented in Table 7. Example segmentation results are shown in Figure 13. The use of HIMs consistently outperformed HFN without HIMs, with an average margin of 1.80% in Jaccard measure across the three datasets.

Figure 13 shows example segmentation results of melanoma studies with and without HIMs. These melanoma studies have inhomogeneous textures. The segmentation results with HIMs are consistent to the appearance and are closer to the ground truth annotations.

Table 7. Segmentation results our HFN method with and without HIMs.

| Datasets | | Jac. | Sen. | Spec. | Acc. |
|---|---|---|---|---|---|
| ISBI 2017 – Overall | HFN (w/o HIMs) | 81.53 | 88.42 | 96.50* | 95.29 |
| | HFN | 83.70* | 92.33* | 96.16 | 95.80* |
| ISBI 2016 – Overall | HFN (w/o HIMs) | 86.77 | 92.58 | 96.68* | 96.37 |
| | HFN | 88.17* | 94.22* | 96.45 | 96.64* |
| PH2 – Overall | HFN (w/o HIMs) | 86.97 | 96.41* | 94.77 | 95.46 |
| | HFN | 88.81* | 95.60 | 95.76* | 96.36* |

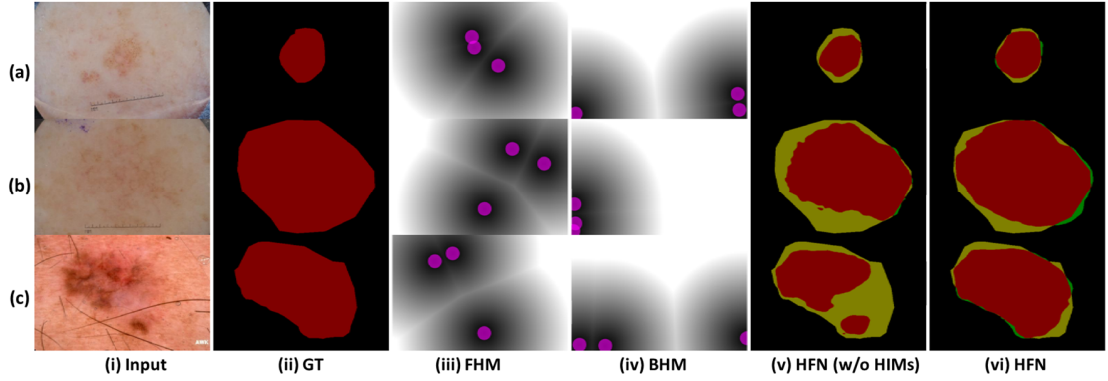

|  | (i) Input | (ii) GT | (iii) FHM | (iv) BHM | (v) HFN (w/o HIMs) | (vi) HFN |

Figure 13. Results for segmenting three melanoma studies of our HFN method with and without HIMs, where (i) input image; (ii) ground truth (GT); (iii, iv) foreground and background hint maps (derived with $\delta(f) = 3$ and $\delta(\beta) = 3$); and (v, vi) segmentation results of our HFN without and with HIMs.

*D. Segmentation Results on Challenging Skin Lesions*

Table 8 shows the segmentation results on the challenging skin lesions. It shows that our HFN improved the fully automatic method DCL with a large margin of 8.62% and a margin of 2.93% to the semi-automatic method P-Net in average in Jaccard measure across the three datasets.

Figure 14 depicts the segmentation results on two challenging skin lesions derived from DCL, P-Net and HFN. These skin lesions present fuzzy boundaries and low-contrast to the background. Both DCL and P-Net failed to depict the skin lesion boundaries. In contrast, our HFN can consistently produce better segmentation results regardless of skin lesion characteristics.

Table 8. Comparison of segmentation results on the challenging studies.

| Datasets | | Jac. | Sen. | Spec. | Acc. |
|---|---|---|---|---|---|
| ISBI 2017 – Overall | DCL | 65.26 | 78.93 | 96.85 | 93.81 |
| | P-Net | 72.09 | 88.42 | 96.82 | 95.79 |
| | HFN | 75.74* | 89.17* | 97.07* | 96.41* |
| ISBI 2016 – Overall | DCL | 71.06 | 90.32 | 91.21 | 91.06 |
| | P-Net | 74.97 | 92.09 | 92.11 | 92.51 |
| | HFN | 78.22* | 92.14* | 94.14* | 93.95* |
| PH2 – Overall | DCL | 76.10 | 96.62* | 92.70 | 93.97 |
| | P-Net | 82.44 | 96.20 | 95.16 | 95.75 |
| | HFN | 84.32* | 96.23 | 96.61* | 96.81* |

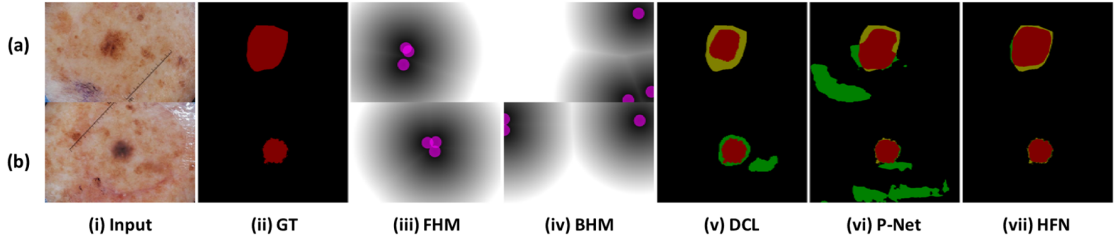

Figure 14. Results for segmenting two challenging skin lesions, where (i) input image; (ii) ground truth (GT); (iii, iv) foreground and background hint maps (derived with $\delta(f) = 3$ and $\delta(\beta) = 3$); and (v, vi, vii) segmentation results of DCL, P-Net and HFN.

*E. Segmentation Results with Noisy User-Inputs*

Table 9 shows the segmentation results with different number of noisy foreground $\delta(f)'$ and background $\delta(\beta)'$ user-inputs. It shows that our HFN has consistent segmentation results even with noisy user-inputs; compared to without using noisy user-inputs, our HFN was 1.44% lower at $(f)' = 1$ and 3.03% lower at $\delta(\beta)' = 1$ in Jaccard measure. We also identified that foreground noisy user-inputs have higher impact to the segmentation accuracy compared to the background counterpart. As expected, the segmentation results were poorer with additional noisy user-inputs. Nevertheless, our method was tolerant to the noisy user-inputs and has competitive segmentation accuracy when compared to the automatic segmentation methods even with extreme noisy user-inputs.

Table 9. Comparison of segmentation results with different number and types of noisy user-inputs, where $\delta(f)'$ and $\delta(\beta)'$ are the number of foreground and background noisy user-inputs.

| ISBI 2017 – Overall | Jac. | Sen. | Spec. | Acc. |
|---|---|---|---|---|
| $\delta(f)' = 0, \delta(\beta)' = 1$ | 82.26 | 88.35 | 97.00 | 95.63 |
| $\delta(f)' = 1, \delta(\beta)' = 0$ | 80.67 | 94.76 | 92.84 | 94.98 |
| $\delta(f)' = 1, \delta(\beta)' = 1$ | 80.25 | 94.67 | 92.81 | 94.87 |
| $\delta(f)' = 2, \delta(\beta)' = 2$ | 77.79 | 96.34 | 90.56 | 94.01 |

## IV. DISCUSSIONS

Our main findings are that: (i) our HFN with user-inputs consistently improved skin lesion segmentation, in particular, for the skin lesions, where the image characteristics are less common in the training datasets; (ii) compared to early-fusion methods, fusing separately extracted complementary features (user-inputs and image features) produced advantages in leveraging user-inputs that resulted in improved segmentation of challenging skin lesions; and (iii) HIMs ensured the appearance of the segmented skin lesions was spatially consistent.

### A. Comparing to Existing Skin Lesion Segmentation Methods on Different Datasets

Our method achieved the best overall performance compared to the existing fully- and semi-automatic segmentation methods on the ISBI 2017, ISBI 2016 and PH2 datasets. The improvements of FCN-8s over the traditional methods such as the SSLS, are due to the FCN-8s being able to encode image-wide semantic information and shallow appearance information that are used in the segmentation. The further improvements from BiDFL, MFCN, DCL, MB and the top 5 performing teams in the competition, was attributed to them using ensembled network architectures and additional training data for supervision (classification labels). All these methods, however, had inconsistent segmentation results across non-melanoma and melanoma studies. They tended to overfit to the dominant non-melanoma skin lesions with a poorer performance on melanoma, which are more difficult to segment when compared to non-melanoma, and also have less training samples. For example, when compared with the recently published BiDFL method on the ISBI 2017 dataset, our method improved the overall segmentation performance of 2.23% in Jaccard measure and a large margin of 3.72% in Jaccard measure for segmenting the more challenging melanoma studies.

Furthermore, existing automatic segmentation methods generated inconsistent segmentation results across different datasets. Compared to the DDN and DAGAN methods on the ISBI 2016 dataset, our HFN method had an improvement of 1.17% and 1.07% in Jaccard measure. Compared to the ISBI 2017, ISBI 2016 dataset has less challenging studies and less variations in lesion locations and illumination. Consequently, when these methods applied to the more challenging ISBI 2017 dataset, these automatic FCN based methods fail to capture all the skin lesion variations especially for the skin lesions that are less common in the training dataset. In contrast, our HFN method leverages user-inputs to learn lesion characteristics and then infer the segmentation results according to the relevance of the user-inputs. As a result, our method had a large improvement of 7.20% and 6.6% in Jaccard measure to the DDN and DAGAN methods on the ISBI 2017 dataset.

Our HFN method has also greatly improved the existing semi-automatic segmentation methods. The improvement of our method to the FCNN, CAGN and P-Net methods is attribute to the use of hyper-fusion instead of early-fusion. Our hyper-fusion network separately processed user-inputs and then fused user-inputs with skin lesion image features across multiple fusion stages, which allowed for the HFN to continuously leverage the user-inputs to segment the skin lesions that are difficult. FCTSFN used a two-stream late fusion network for segmentation, where the images and the user-inputs were separately processed by two FCN networks with the resultant features fused. However, the late fusion of extracted image features tends to dismiss the correlations between the image and the user-inputs and the correlations may only accessible at the early stage of the network. In addition, FCTSFN was designed for natural image segmentation and the segmentation performance was reliant on using an image classification network trained with a large natural image dataset. For skin lesion segmentation, there was no equivalent skin lesion classification network model that can be adapted for initialization. Consequently, FCTSFN failed to provide accurate segmentation for skin lesions.

*B. Analysis of User-Inputs*

P-Net was the second best performing semi-automatic segmentation method. As expected, the segmentation results for both P-Net and our HFN were improved with additional user-inputs. However, with P-Net, as an early-fusion method, the first convolutional layer was used to combine both the skin lesion images and the user-input hint maps, and then to derive the fused image features. Therefore, the user-inputs information could potentially be lost after the first convolutional layer. Consequently, P-Net had limited segmentation improvement with additional user-inputs and had difficulty in segmenting the challenging skin lesions e.g., inhomogeneous textures, fuzzy boundaries. In contrast, our HFN separately processed user-inputs and then fused user-inputs with skin lesion image features across multiple fusion stages. Multiple fusion stages have the advantages in leveraging user-inputs to iteratively refine the segmentation results, which further improved the segmentation results on the challenging skin lesions, as shown in Fig. 15. Fig. 15 presents feature visualization at each fusion stages of two challenging skin lesions. It shows that by leveraging the user-inputs, HFN was capable of gradually focusing on the skin lesions that are difficult to be captured automatically.

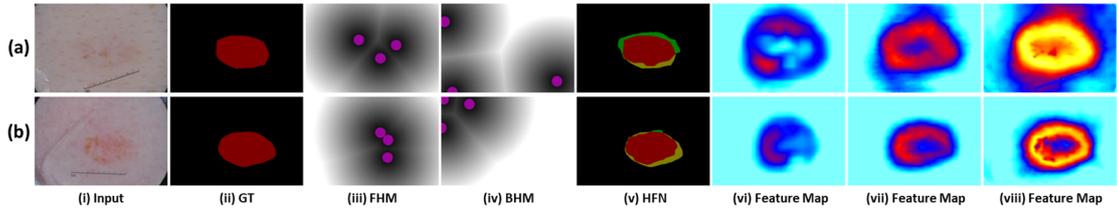

Figure 15. Feature visualization of two challenging skin lesions, where (i) input image; (ii) ground truth (GT); (iii, iv) foreground and background hint maps (derived with $\delta(f) = 3$ and $\delta(\beta) = 3$); (v) segmentation results of HFN, (v, vi, vii) are feature map visualization results of HFN at stages 1 to 3 on the decoder side. Blue color indicates low weight; yellow and red indicate higher weight.

### C.  Analysis of Individual Components

The HIMs improved the overall segmentation performance across all three datasets. We explain this by the HIMs enabling user-inputs to guide and constrain the learning of skin lesion characteristics and then to propagate the intermediary segmentation results to the next stage. The guidance and constraint ensured the appearance of the segmented skin lesions was in agreement with the skin lesion images and its derived hint maps, and this enabled the segmentation of skin lesions to be spatially consistent.

### D.  Future Work

#### 1)  Application to Total-Body 3-Dimensional (3D) Photography

Total-body 3D photography, currently being implemented in the clinic, that constructs a digital 3D avatar of the patient that can be used to view and monitor skin lesions across the body over time. When compared to current manual dermoscopy and limited-access time-consuming 2D total body photography systems, total-body 3D photography brings new spatial and temporal capabilities and skin lesions at different sites of the body and at different times can be detected simultaneously. The Australian Centre of Excellence in Melanoma Imaging and Diagnosis and Memorial Sloan Kettering Cancer Center have recently installed this new technology. The massive amounts of data collected from the total-body 3D photography systems offer different challenges to tedious manual interpretation process of skin lesion.

Our intention is to explore the adaption of our method to total-body 3D photography. We suggest that our method, coupled with incremental learning techniques, and with minimum amounts of user inputs will be sufficient to accurately segment skin lesions across different sites and time points. Further, the large annotated International Skin Imaging Collaboration (ISIC) archive has facilitated the development of numerous deep learning based skin lesion analysis algorithms. Despite new imaging capabilities from the total-body 3D photography, the development of large annotated datasets for total-body 3D photography systems has not kept pace. Based on the superior generalizability of the proposed method, we plan to explore the adaptions of our method trained with ISIC dataset to the images collected from total-body 3D photography systems. We believe this will facilitate to curate a large total-body 3D photography dataset that can be shared to the public and to increase research capacity in the community.

*2) Evaluation of Different Types of User-Inputs*

For this work, we mainly considered user-clicks as the user-inputs for segmentation. Alternative user-inputs such as bounding boxes and scribbles are other possible approaches. Bounding boxes use a rectangular border to fully enclose the skin lesion and scribbles use rough drawings to indicate the foreground and background skin lesions. In our future work, we plan to evaluate our approach with these types of user-inputs and to evaluate their performance differences e.g., segmentation accuracy and efficiency and tolerance to noisy user-inputs.

## V. CONCLUSIONS

In this paper, we proposed a method to segment skin lesions in a semi-automated manner. Our method used a deep hyper-fusion FCN to iteratively fuse, separately extracted user-input features, with skin lesion image features and to continuously leverage user-inputs to guide and constrain the learning of skin lesion characteristics. By learning and inferring user-inputs derived from few user-clicks, we achieved accurate segmentation results for skin lesions that are known to be challenging, such as those with fuzzy boundaries, inhomogeneous textures and low-contrast to the background. Our HFN had consistently better segmentation results on three well-established public datasets (ISBI 2017, ISBI 2016 Skin Lesion Challenge datasets and PH2 dataset) and this suggests that the HFN is generalizable and it can also be applied to skin lesion images acquired with different scanning protocols e.g., field-of-view.

## VI. Acknowledgement

This work was supported in part by Australia Research Council (ARC) grants (IC170100022 and DP200103748).